\documentclass{article}
\usepackage{graphicx} 
\usepackage[nonatbib,final]{bdu_2024}
\usepackage{amsmath}
\usepackage{amssymb}
\usepackage{biblatex}
\usepackage{algorithm}
\usepackage{algpseudocodex}
\usepackage{enumitem}
\usepackage{mathrsfs}
\usepackage{booktabs}
\usepackage{biblatex}
\DeclareMathOperator*{\argmax}{argmax}
\DeclareMathOperator*{\argmin}{argmin}
\newcommand{\dd}{\mathop{}\!{d}}

\title{Fast, Precise Thompson Sampling for Bayesian Optimization}
\author{
David Sweet\\
Department of Computer Science\\
Yeshiva University\\
New York, NY 10174\\
\texttt{david.sweet@yu.edu}
}
\date{August 2024}

\begin{document}

\maketitle

\begin{abstract}
Thompson sampling (TS) has optimal regret and excellent empirical performance in multi-armed bandit problems. Yet, in Bayesian optimization, TS underperforms popular acquisition functions (e.g., EI, UCB). TS samples arms according to the probability that they are optimal. A recent algorithm, P-Star Sampler (PSS), performs such a sampling via Hit-and-Run. We present an improved version, Stagger Thompson Sampler (STS). STS more precisely locates the maximizer than does TS using less computation time. We demonstrate that STS outperforms TS, PSS, and other acquisition methods in numerical experiments of optimizations of several test functions across a broad range of dimension. Additionally, since PSS was originally presented not as a standalone acquisition method but as an input to a batching algorithm called Minimal Terminal Variance (MTV), we also demonstrate that STS matches PSS performance when used as the input to MTV.
\end{abstract}

\section{Introduction}
Bayesian optimization (BO) is applied to many experiment- and simulation-based optimization problems in science and engineering \cite{mtv}. The aim of BO methods is to minimize the number of measurements needed to find a good system configuration. Measurements are taken in a sequence of batches of one or more \textit{arms} -- where an arm is one system configuration. System performance is measured for each arm in a batch, then a new batch is produced, performance is measured, and so on.

Thompson sampling (TS) samples arms according to the probability that they maximize system performance \cite{ts0}. Let's denote an arm as $x_a$, its performance as $f(x_a)$, and the probability that an arm is best, i.e., that $x_a = \argmax_{x} f(x)$, as $p_*(x)$. TS draws an arm as: $x_a \sim p_*(x)$. TS has optimal or near-optimal regret \cite{agrawal2012optimalregretboundsthompson,NIPS2011_e53a0a29} for multi-armed bandits (MAB) and is also used in BO \cite{ts}. 

Application of TS to BO is not straightforward. Since BO arms are typically continuous -- e.g., $x \in [0,1]^d$ -- sampling from $p_*(x)$ is non-trivial. The usual approach is to first sample many candidate arms uniformly, $x_i \sim \mathcal{U}([0,1]^d)$, then draw a value, $y_i$, from a model distribution of $f(x)$ at each $x_i$. The $x_i$ that yields the highest-valued draw, i.e., $x_a = \argmax_{x_i} y(x_i)$, is taken as the arm. BO methods usually model $y(x)$ with a Gaussian process, $\mathcal{GP}$ \cite{gprbook}, i.e. $y(x) \sim \mathcal{N}(\mu(x), \sigma^2(x))$. While appealing for its simplicity, TS, implemented as described, tends to underperform popular acquisition functions such as Expected Improvement (EI) \cite{eiopt} or Upper Confidence bound (UCB) \cite{ucbbo} (see Subsection~\ref{subsec:oapb}).

\section{Stagger Thompson Sampling}

The TS arm candidates, being uniform in $[0,1]^d$, are unlikely to fall where $p_*(x)$ has high density. If we imagine that the bulk of $p_*(x)$ lies in a hypercube of side $\varepsilon < 1$, then the probability that a randomly-chosen $x \in [0,1]^d$ falls in the hypercube is just the hypercube volume, $v = \varepsilon^d$. Note that (i) $v$ decreases exponentially with dimension, and (ii) $\varepsilon$, thus $v$, decreases with each additional measurement added to the model of $p_*(x)$ (see figure~\ref{fig:tststs}(c) in appendix~\ref{app:spt}). 
Effect (ii) is the aim of BO -- to localize the maximizer. Effect (i) is the "curse of dimensionality". Thus, the number of candidates required to find the bulk of $p_*(x)$ (i) increases exponentially in dimension, and (ii) increases with each additional measurement (albeit in a non-obvious way).

The P-Star Sampler (PSS) \cite{mtv} is a Hit-and-Run \cite{hnr} sampler with a Metropolis filter \cite{mcmc}. Our algorithm, Stagger Thompson Sampler (STS), is, also, but differs in several details, discussed below algorithm~\ref{alg:stagger}.

\begin{algorithm}
\caption{Stagger Thompson Sampler}
\label{alg:stagger}
\renewcommand{\thealgorithm}{}
\floatname{algorithm}{}
\begin{algorithmic}[1]
    \If{no measurements yet}
    \State \Return $x_i \sim \mathcal{U}([0,1]^d)$ \Comment{Take $p_*(x)$ prior as uniform in $x$}
    \EndIf
    \State $\tilde{x}_* = \argmax_{x} \mu(x)$ \Comment{$\mu(x)$ is mean of a given $\mathcal{GP}$}
    \State $x_a  = \tilde{x}_*$ \Comment{Initialize arm}
    \ForAll{$m \in 1, \dots, M$} \Comment{Refine  arm $M$ times}
    \State $x_t = \mathcal{U}([0,1]^d)$ \Comment{Perturbation target}
    \State $s = e^{-k\mathcal{U}([0, 1])}$ \Comment{A "stagger" perturbation length}
    \State $x_a^{\prime}= x_a  + s (x_t - x_a )$ \Comment{Perturbations}
    \State $[ y, y^\prime ] \sim \mathcal{GP}([x_a,x_a^\prime])$  \Comment{Joint sample}
    \State $x_a \leftarrow x_a^\prime$ \textbf{if} $y^\prime > y$ \Comment{MH acceptance}
    \EndFor
    \State \Return $x_a$ \Comment{A sample from $p_*(x)$}

\end{algorithmic}\end{algorithm}

Algorithm~\ref{alg:stagger} modifies vanilla Hit-and-Run in two ways: (i) Instead of initializing randomly, we initialize an arm candidate, $x_a$, at $\tilde{x}_*$. (ii) Instead of perturbing uniformly along some direction -- since the scale of $p_*(x)$ is unknown and may be small -- we choose the length of the perturbation uniformly in its exponent, i.e. as $\sim e^{-kU}$, a log-uniform random variable. ($k$ is a hyperparameter, which we choose to be $k=\ln 10^{-6}$). Numerical ablation studies in appendix~\ref{app:ablation} show that these modifications improve performance in optimization. We refer to the log-uniform perturbation as a "stagger" proposal, following \cite{stagger}. Besides being empirically effective, a stagger proposal also obviates the need to adapt the scale of the proposal distribution, as is done in PSS (which uses a Gaussian proposal). We see this as a valuable, practical simplification.

Since a log-uniform distribution is a symmetric proposal we expect the Markov chain generated by algorithm~\ref{alg:stagger} to converge to $p_*(x)$. Appendix~\ref{app:spt} provides some numerical support for this. 

Perturbations are made along a line from $x_a$ to a target point, $x_t$, which is chosen uniformly inside the bounding box, $[0,1]^d$. This ensures that the final perturbation [a convex combination of points in the (convex) bounding box] will lie inside the bounding box. It also simplifies the implementation somewhat, since boundary detection is unnecessary. PSS performs a bisection search to find the boundary of the box along a randomly-oriented line passing through $x_a$.

\begin{figure}[ht]
   \centering
   \includegraphics[width=0.75\linewidth]{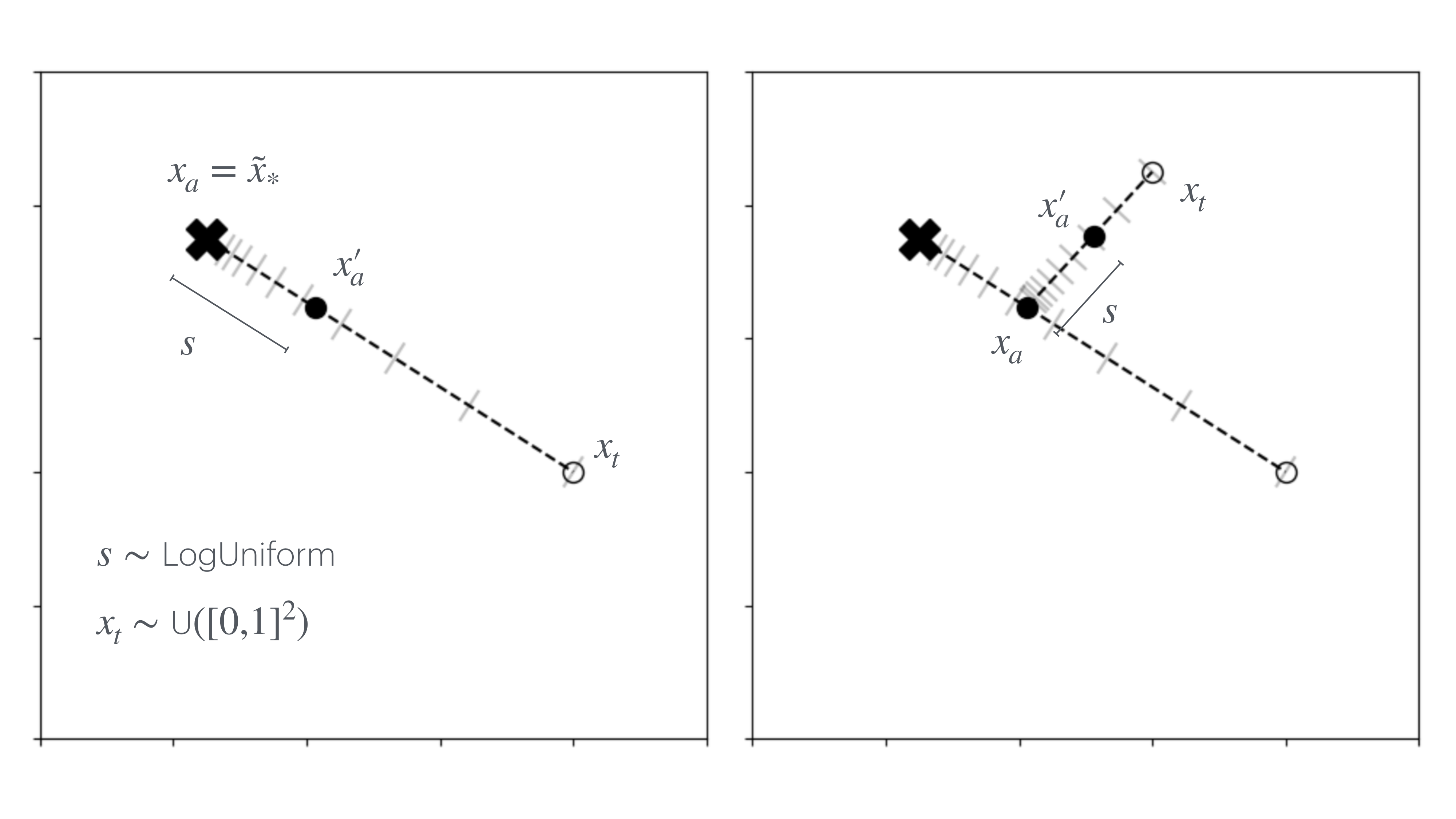}\hfill
    \caption{Two iterations of the for loop in algorithm~\ref{alg:stagger}. Hash marks indicate the log-uniform (stagger) distribution for $s$. A Thompson sample -- a joint sample, $\mathcal{GP}([x_a,x_a^\prime])$ --   determines whether $x_a$ updates to $x_a^\prime$.
    }
    \label{fig:sts_algo}
\end{figure}

We accept a perturbation of $x$, $x^\prime$, with Metropolis acceptance probability $p_{acc} = \min \{1, p_*(x^\prime)/p_*(x)\}$. As a coarse (and fast) approximation to $p_{acc}$, we follow PSS and just take a single joint sample from $\mathcal{GP}$ and accept whichever point, $x$ or $x^\prime$, has a larger sample value. Note that this is a Thompson sample from the set $\{x,x^\prime\}$, so we might describe STS as iterated Thompson sampling.

Appendix~\ref{app:spt} offers numerical evidence that (i) samples from STS are nearer the true maximizer than are samples from TS, and (ii) STS produces samples more quickly than standard TS while better approximating $p_*(x)$.

Previous work applying MCMC methods to Thompson sampling include random-walk Metropolis algorithms constrained to a trust region \cite{mcmcbo} and a sequential Monte Carlo algorithm \cite{smcbo}.

\section{Numerical Experiments}

To evaluate STS, we optimize various test functions, tracking the maximum measured function value at each round and comparing the values to those found by other methods. We use the term \textit{round} to refer to the generation of one or more arms followed by the measurement of them.

\subsection{Ackley-200d}
\label{sec:ackley}

\begin{figure}[ht]
   \centering
   \includegraphics[width=0.5\linewidth]{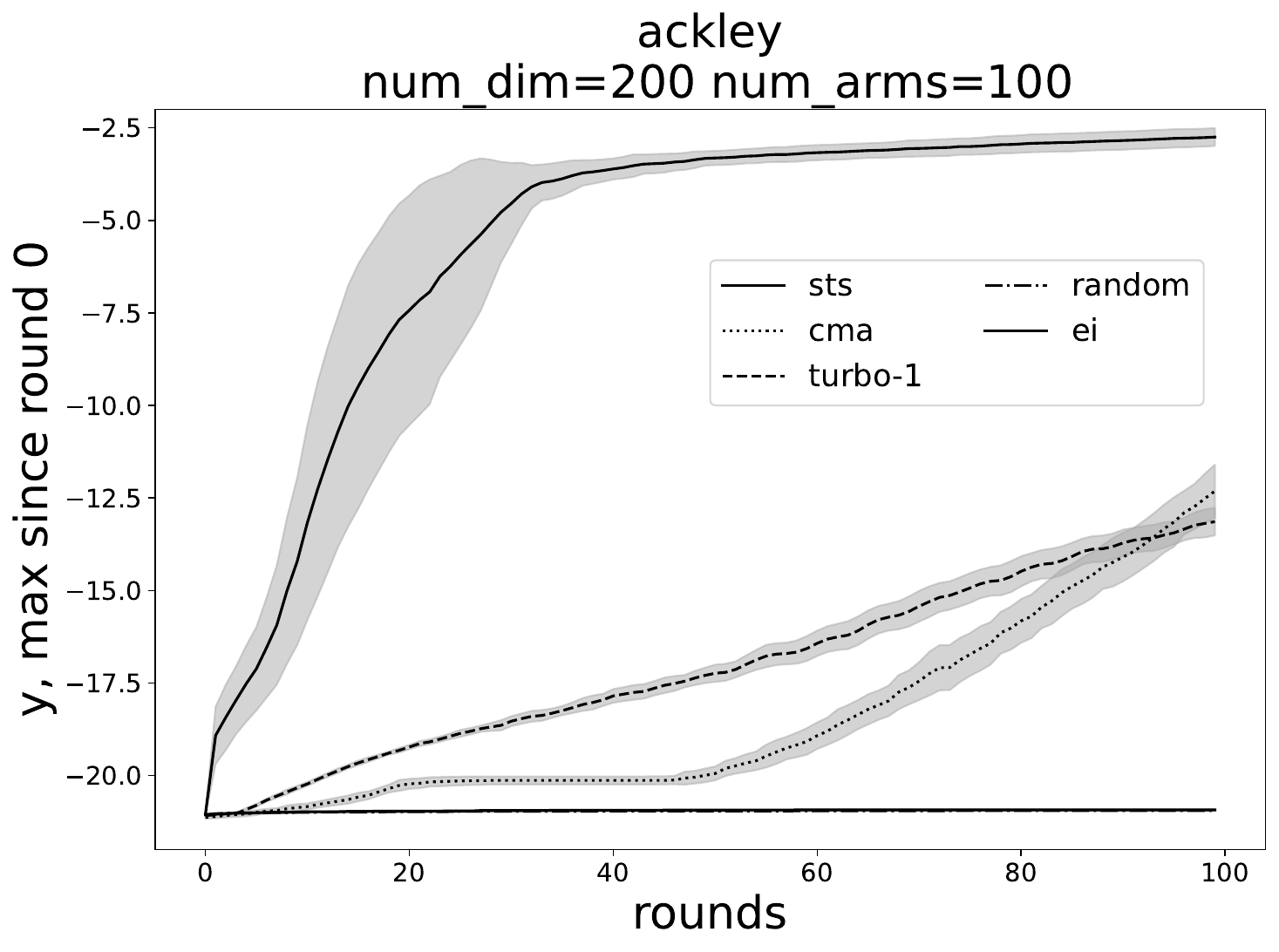}\hfill
    \caption{We maximize the Ackley function in 200 dimensions over 100 rounds of 100 arms/round. The error areas are twice the standard error over 10 runs. STS (\texttt{sts}) finds higher values more quickly than other optimization methods: \texttt{turbo-1} - TuRBO \cite{turbo} with one trust region. \texttt{cma} - CMA-ES \cite{cmaes}, an evolution strategy. \texttt{random} - Choose arms uniformly randomly (serving as a baseline). \cite{turbo}. 
    }
    \label{fig:ackley200d}
\end{figure}

To introduce our comparison methodology, we compare STS to a few other optimization methods, in particular to TuRBO, a trust-region-enhanced Thompson sampling method \cite{turbo}.  In \cite{turbo}, the authors optimize the Ackley function in 200 dimensions with 100 arms/round on a restricted subspace of parameters. Figure~\ref{fig:ackley200d} optimizes the same function on the standard parameter space using STS, TuRBO, and other methods. STS finds higher values of $y$ more quickly than the other methods.

We can summarize each method's performance in Figure~\ref{fig:ackley200d} with a single number, which we'll call the \textit{score}. At each round, $i$, find the maximum measured values so far for each method, $m$: $y_{i,m}$. Rank these values across $m$ and scale: $r_{i,m} = [\texttt{rank}(y_{i,m})-1]/(M-1)$, where $M$ is the number of methods. Repeat this for every round, $i$, then average over all $R$ rounds to get the score: $s_m = \sum_i^R{r_{i,m}}/R$. The scores in figure~\ref{fig:ackley200d} are $s_{\texttt{sts}}=1$, $s_{\texttt{turbo-1}}=2/3$, $s_{\texttt{cma}}=1/2$, and $s_{\texttt{random}}=0$.
Using an normalized score enables us to average over runs on different functions (which, in general, have different scales for $y$). Using a rank-based score prevents a dramatic result, like the one in figure~\ref{fig:ackley200d}, from dominating the average.

In our experiments below we optimize over nine common functions. To add variety to the function set and to avoid an artifact where an optimization method might coincidentally prefer to select points near a function's optimum (e.g., at the center of the parameter space), we randomly distort each function as in \cite{mtv}, repeating the optimization 30 times with different random distortions.

\subsection{One arm per round}
\label{subsec:oapb}

\begin{figure}[ht]
   \centering
   \includegraphics[width=0.75\linewidth]{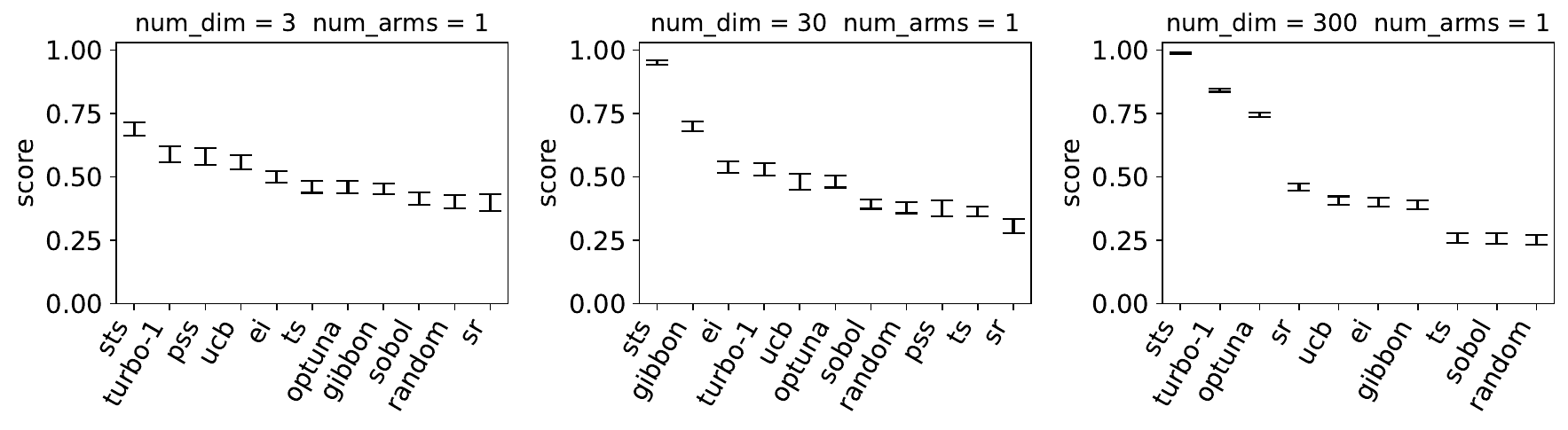}\hfill
    \caption{
    We optimize for $\max(30, \texttt{num\_dim})$ rounds with \texttt{num\_arms} / round over the functions ackley, dixonprice, griewank, levy, michalewicz, rastrigin, rosenbrock, sphere, and stybtang \cite{opttest} with random distortions (see section~\ref{sec:ackley}).  Error bars are two standard errors over all functions and 30 runs/function. Figure represents a total of 874,800 function evaluations. (We were not able to run \texttt{pss} for \texttt{num\_dim}=300 due to long computation times.
    }
    \label{fig:compare}
\end{figure}

We compare STS to other BO methods in dimensions 3 through 300, all generating one arm per round. For each dimension, each method's score is averaged over all test functions. See Figure~\ref{fig:compare}. STS has the highest score in each dimension, and its advantage appears to increase with dimension. Data from dimensions 1, 10, and 100 (unpublished for space) follow the same pattern.

The various optimization methods are: \texttt{sts} - Stagger Thompson Sampling (Algorithm~\ref{alg:stagger}), \texttt{random} - Uniformly-random arm, \texttt{sobol} - Uniform, space-filling arms \cite[Chapter~5]{dace}, \texttt{sr} - Simple Regret, $\mu(x)$, \texttt{ts} - Thompson Sampling \cite{ts}, \texttt{ucb} - Upper Confidence Bound \cite{ucbbo,q}, \texttt{ei} - Expected Improvement \cite{mockus,q}, \texttt{gibbon} - GIBBON,  an entropy-based method \cite{gibbon}, and \texttt{optuna} - an open-source optimizer \cite{optuna} employing a tree-structured Parzen estimator \cite{tpe}. For the methods \texttt{ei}, \texttt{lei}, \texttt{ucb}, \texttt{sr}, \texttt{gibbon}, and \texttt{turbo-1}, we initialize by taking a Sobol' sample for the first round's arm. \texttt{sts} and \texttt{ts} do not require initialization.

STS makes no explicit accommodations for higher-dimensional problems yet performs well on them. Of the methods evaluated, only \texttt{turbo-1} specifically targets higher-dimensional problems \cite{turbo}, so it may be valuable for future work to compare STS to other methods specifically designed for such problems. (See references to methods in \cite{turbo}.)

\subsection{Multiple arms per round}

Thompson sampling can be extended to batches of more than one arm simply by taking multiple samples from $p_*(x)$, e.g., by running Algorithm~\ref{alg:stagger} multiple times. However, this approach can be inefficient \cite{dpp} because some samples -- since they are independently generated -- may lie very near each other and, thus, provide less information about $f(x)$ than if they were to lie farther apart. This problem, that of generating effective batches of arms, is not unique to TS but exists for all approaches to acquisition, and there are various methods for dealing with it \cite{dpp} \cite{q} \cite{gibbon}.

One method, Minimal Terminal Variance (MTV) \cite{mtv}, minimizes the post-measurement, average variance of the GP, weighted by $p_*(x)$:
\begin{equation}
\label{eq:int}
MTV(x_a) = \int \dd{x} \ p_*(x) \sigma^2(x|x_a)
\end{equation}
approximated by $\sum_{i} \sigma^2(x_i|x_a)$, where $x_i$ are drawn from $x_i \sim p_*(x)$ with P-Star Sampler (PSS). MTV is interesting, in part, because it can design experiments both when prior measurements are available and \textit{ab initio} (e.g., at initialization time). It not only outperforms acquisition functions (like EI or UCB) but the same formulation also outperforms common initialization methods, such as Sobol' sampling \cite{mtv}. 

We modify MTV to draw $x_i \sim p_*(x)$ using STS instead of PSS. Note, also, that the arms, $x_a$, that minimize $MTV(x_a)$ are not drawn from the set $x_i$ but are chosen by a continuous minimization algorithm (specifically, $\texttt{scipy.minimize}$, as implemented in \cite{botorch_code}), such that $x_a = \argmin_{x_a^\prime}\sum_{i} \sigma^2(x_i|x_a^\prime)$.

\begin{figure}[ht]
   \centering
   \includegraphics[width=0.9\linewidth]{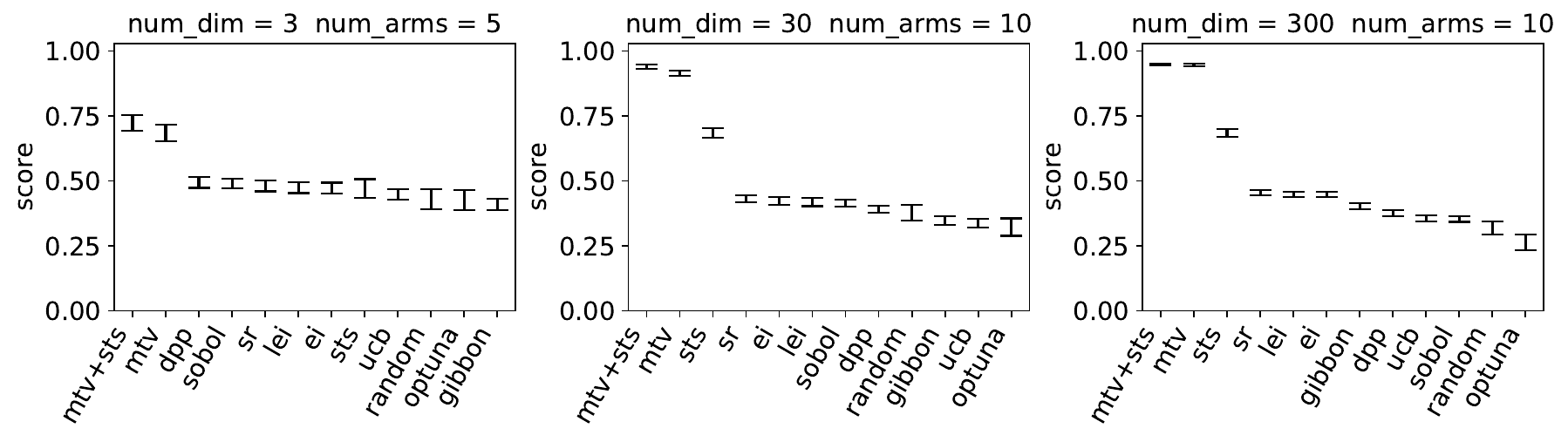}\hfill
    \caption{Optimizations with 3 multi-arm rounds on nine test functions. MTV+STS (\texttt{mtv+sts}) outperforms all other methods across a range of dimensions. The figure consists of $1.2 \cdot 10^6$ function evaluations. Calculations (not shown) for 1, 10, and 100 dimensions show similar results.}
    \label{fig:batches}
\end{figure}

Figure~\ref{fig:batches} compares MTV, with P-Star Sampler replaced by STS (\texttt{mtv+sts}), to other methods using various dimensions (\texttt{num\_dim}) and batch sizes (\texttt{num\_arms}): \texttt{mtv} - MTV, as in \cite{mtv},  \texttt{lei} - q-Log EI, an improved EI \cite{logei}, and \texttt{dpp} - DPP-TS \cite{dpp}, a diversified-batching TS. For the methods \texttt{ei}, \texttt{ucb}, \texttt{sr}, \texttt{gibbon}, and \texttt{dpp}, we initialize by taking Sobol' samples for the first round. \texttt{sts}, \texttt{mtv}, and \texttt{mtv+sts} do not require initialization.

Figure~\ref{fig:batches} roughly reproduces figure 3 of \cite{mtv}, adding more methods and extending to higher dimensions. Additionally, we include pure PSS and STS sampling, where arms are simply independent draws from $p_*(x)$, to highlight the positive impact of MTV on batch design.

When MTV's input samples come from STS (\texttt{mtv+sts}), performance is similar to the original MTV (\texttt{mtv}).

\section{Conclusion}

We presented Stagger Thompson Sampler, which is novel in several ways:
\begin{itemize}
    \item It outperforms not only the standard approach to Thompson sampling but, also, popular acquisition functions, a trust region Thompson sampling method, and an evolution strategy.
    \item It is simpler and more effective at acquisition than an earlier sampler (P-Star Sampler).
    \item It works on high-dimensional problems without modification.
\end{itemize}
Additionally, the combination MTV+STS is unique in that it applies to so broad a range of optimization problems: It solves problems with zero or more pre-existing measurements, with one or more arms/batch, and in dimensions ranging from low to high.

\acksection
This work was carried out in affiliation with Yeshiva University. The author is additionally affiliated with DRW Holdings, LLC. This work was supported, in part, by a grant from Modal.

\clearpage
\printbibliography

@misc{mtv,
      title={Optimal Initialization of Batch Bayesian Optimization}, 
      author={Jiuge Ren and David Sweet},
      year={2024},
      eprint={2404.17997},
      archivePrefix={arXiv},
      primaryClass={cs.LG},
      url={https://arxiv.org/abs/2404.17997}, 
}

@misc{agrawal2012optimalregretboundsthompson,
      title={Further Optimal Regret Bounds for Thompson Sampling}, 
      author={Shipra Agrawal and Navin Goyal},
      year={2012},
      eprint={1209.3353},
      archivePrefix={arXiv},
      primaryClass={cs.LG},
      url={https://arxiv.org/abs/1209.3353}, 
}

@inproceedings{NIPS2011_e53a0a29,
 author = {Chapelle, Olivier and Li, Lihong},
 booktitle = {Advances in Neural Information Processing Systems},
 editor = {J. Shawe-Taylor and R. Zemel and P. Bartlett and F. Pereira and K.Q. Weinberger},
 pages = {},
 publisher = {Curran Associates, Inc.},
 title = {An Empirical Evaluation of Thompson Sampling},
 url = {https://proceedings.neurips.cc/paper_files/paper/2011/file/e53a0a2978c28872a4505bdb51db06dc-Paper.pdf},
 volume = {24},
 year = {2011}
}

@InProceedings{ts,
  title = 	 {Parallelised Bayesian Optimisation via Thompson Sampling},
  author = 	 {Kandasamy, Kirthevasan and Krishnamurthy, Akshay and Schneider, Jeff and Poczos, Barnabas},
  booktitle = 	 {Proceedings of the Twenty-First International Conference on Artificial Intelligence and Statistics},
  pages = 	 {133--142},
  year = 	 {2018},
  editor = 	 {Storkey, Amos and Perez-Cruz, Fernando},
  volume = 	 {84},
  series = 	 {Proceedings of Machine Learning Research},
  month = 	 {4},
  publisher =    {PMLR},
  url = 	 {https://proceedings.mlr.press/v84/kandasamy18a.html}
}

@InProceedings{mockus,
author="Mo{\v{c}}kus, J.",
editor="Marchuk, G. I.",
title="On bayesian methods for seeking the extremum",
booktitle="Optimization Techniques IFIP Technical Conference Novosibirsk, July 1--7, 1974",
year="1975",
publisher="Springer Berlin Heidelberg",
address="Berlin, Heidelberg",
pages="400--404",
isbn="978-3-540-37497-8"
}

@misc{q,
      title={The reparameterization trick for acquisition functions}, 
      author={James T. Wilson and Riccardo Moriconi and Frank Hutter and Marc Peter Deisenroth},
      year={2017},
      eprint={1712.00424},
      archivePrefix={arXiv},
      primaryClass={stat.ML},
      url={https://arxiv.org/abs/1712.00424}, 
}

@misc{ucbbo,
      title={Parallelizing Exploration-Exploitation Tradeoffs with Gaussian Process Bandit Optimization}, 
      author={Thomas Desautels and Andreas Krause and Joel Burdick},
      year={2012},
      eprint={1206.6402},
      archivePrefix={arXiv},
      primaryClass={cs.LG},
      url={https://arxiv.org/abs/1206.6402}, 
}

@article{stagger,
  title = {Stagger-and-Step Method: Detecting and Computing Chaotic Saddles in Higher Dimensions},
  author = {Sweet, David and Nusse, Helena E. and Yorke, James A.},
  journal = {Phys. Rev. Lett.},
  volume = {86},
  issue = {11},
  pages = {2261--2264},
  numpages = {0},
  year = {2001},
  month = {3},
  publisher = {American Physical Society},
  doi = {10.1103/PhysRevLett.86.2261},
  url = {https://link.aps.org/doi/10.1103/PhysRevLett.86.2261}
}

@book{mcmc,
  author = {Gilks, W.R. and Richardson, S. and Spiegelhalter, D.},
  isbn = {9780412055515},
  lccn = {98033429},
  publisher = {Taylor \& Francis},
  series = {Chapman \& Hall/CRC Interdisciplinary Statistics},
  title = {Markov Chain Monte Carlo in Practice},
  url = {http://books.google.com/books?id=TRXrMWY\_i2IC},
  year = 1995
}

@inproceedings{turbo,
 author = {Eriksson, David and Pearce, Michael and Gardner, Jacob and Turner, Ryan D and Poloczek, Matthias},
 booktitle = {Advances in Neural Information Processing Systems},
 editor = {H. Wallach and H. Larochelle and A. Beygelzimer and F. d\textquotesingle Alch\'{e}-Buc and E. Fox and R. Garnett},
 pages = {},
 publisher = {Curran Associates, Inc.},
 title = {Scalable Global Optimization via Local Bayesian Optimization},
 url = {https://proceedings.neurips.cc/paper_files/paper/2019/file/6c990b7aca7bc7058f5e98ea909e924b-Paper.pdf},
 volume = {32},
 year = {2019}
}

@misc{cmaes,
      title={The CMA Evolution Strategy: A Tutorial}, 
      author={Nikolaus Hansen},
      year={2023},
      eprint={1604.00772},
      archivePrefix={arXiv},
      primaryClass={cs.LG},
      url={https://arxiv.org/abs/1604.00772}, 
}

@misc{opttest,
  author = {S. Surjanovic and D. Bingham},
  title = {Virtual Library of Simulation Experiments: Test Functions and Datasets},
  url = {https://www.sfu.ca/~ssurjano/optimization.html},
  note = {Accessed: 2024-08-20},
  year = {2024}
}

@book{dace,
author = {Thomas J. Santner and Brian J. Williams and William I. Notz},
publisher = {Springer New York, NY},
isbn = {9781493988471},
title = {The Design and Analysis of Computer Experiments},
booktitle = {The Design and Analysis of Computer Experiments},
doi = {https://doi.org/10.1007/978-1-4939-8847-1},
url = {https://link.springer.com/book/10.1007/978-1-4939-8847-1},
year = {2019},
}

@article{gibbon,
  author  = {Henry B. Moss and David S. Leslie and Javier Gonzalez and Paul Rayson},
  title   = {GIBBON: General-purpose Information-Based Bayesian Optimisation},
  journal = {Journal of Machine Learning Research},
  year    = {2021},
  volume  = {22},
  number  = {235},
  pages   = {1--49},
  url     = {http://jmlr.org/papers/v22/21-0120.html}
}

@InProceedings{dpp,
  title = 	 { Diversified Sampling for Batched Bayesian Optimization with Determinantal Point Processes },
  author =       {Nava, Elvis and Mutny, Mojmir and Krause, Andreas},
  booktitle = 	 {Proceedings of The 25th International Conference on Artificial Intelligence and Statistics},
  pages = 	 {7031--7054},
  year = 	 {2022},
  editor = 	 {Camps-Valls, Gustau and Ruiz, Francisco J. R. and Valera, Isabel},
  volume = 	 {151},
  series = 	 {Proceedings of Machine Learning Research},
  month = 	 {3},
  publisher =    {PMLR},
  url = 	 {https://proceedings.mlr.press/v151/nava22a.html},
}

@book{gprbook,
  author = {Rasmussen, Carl Edward and Williams, Christopher K. I.},
  ee = {https://www.worldcat.org/oclc/61285753},
  isbn = {026218253X},
  pages = {I-XVIII, 1-248},
  publisher = {MIT Press},
  series = {Adaptive computation and machine learning},
  title = {Gaussian processes for machine learning.},
  year = 2006
}

@inproceedings{logei,
author = {Daulton, Samuel and Ament, Sebastian and Eriksson, David and Balandat, Maximilian and Bakshy, Eytan},
title = {Unexpected improvements to expected improvement for Bayesian optimization},
year = {2024},
publisher = {Curran Associates Inc.},
address = {Red Hook, NY, USA},
booktitle = {Proceedings of the 37th International Conference on Neural Information Processing Systems},
articleno = {904},
numpages = {36},
location = {New Orleans, LA, USA},
series = {NIPS '23}
}

@article{eiopt,
author = {Li, Yiou and Deng, Xinwei},
title = {An efficient algorithm for Elastic I-optimal design of generalized linear models},
journal = {Canadian Journal of Statistics},
volume = {49},
number = {2},
pages = {438-470},
doi = {https://doi.org/10.1002/cjs.11571},
url = {https://onlinelibrary.wiley.com/doi/abs/10.1002/cjs.11571},
eprint = {https://onlinelibrary.wiley.com/doi/pdf/10.1002/cjs.11571},
year = {2021}
}

@article{ts0,
    author = {William R Thompson},
    title = {On the likelihood that one unknown probability exceeds another in view of the evidence of two samples},
    journal = {Biometrika},
    volume = {25},
    number = {3-4},
    pages = {285-294},
    year = {1933},
    month = {12},
    issn = {0006-3444},
    doi = {10.1093/biomet/25.3-4.285},
    url = {https://doi.org/10.1093/biomet/25.3-4.285},
    eprint = {https://academic.oup.com/biomet/article-pdf/25/3-4/285/513725/25-3-4-285.pdf},
}

@misc{botorch_code,
  author = {Meta},
  title = {BoTorch},
  url = {https://botorch.org},
  year = {2024}
}

@misc{optuna,
  author = {Optuna},
  title = {Optuna},
  url = {https://optuna.org},
  year = {2024}
}

@inproceedings{hnr,
author = {Smith, Robert L.},
title = {The hit-and-run sampler: a globally reaching Markov chain sampler for generating arbitrary multivariate distributions},
year = {1996},
isbn = {0780333837},
publisher = {IEEE Computer Society},
address = {USA},
url = {https://doi.org/10.1145/256562.256619},
doi = {10.1145/256562.256619},
booktitle = {Proceedings of the 28th Conference on Winter Simulation},
pages = {260–264},
numpages = {5},
location = {Coronado, California, USA},
series = {WSC '96}
}

@misc{mcmcbo,
      title={Improving sample efficiency of high dimensional Bayesian optimization with MCMC}, 
      author={Zeji Yi and Yunyue Wei and Chu Xin Cheng and Kaibo He and Yanan Sui},
      year={2024},
      eprint={2401.02650},
      archivePrefix={arXiv},
      primaryClass={cs.LG},
      url={https://arxiv.org/abs/2401.02650}, 
}

@misc{smcbo,
      title={A sequential Monte Carlo approach to Thompson sampling for Bayesian optimization}, 
      author={Hildo Bijl and Thomas B. Schön and Jan-Willem van Wingerden and Michel Verhaegen},
      year={2017},
      eprint={1604.00169},
      archivePrefix={arXiv},
      primaryClass={stat.ML},
      url={https://arxiv.org/abs/1604.00169}, 
}

@article{tpe,
  title   = {Tree-structured {P}arzen estimator: Understanding its algorithm components and their roles for better empirical performance},
  author  = {S. Watanabe},
  journal = {arXiv preprint arXiv:2304.11127},
  year    = {2023}
}

\clearpage
\appendix

\section{Speed, Precision, and Thompson Sampling}
\label{app:spt}

In this appendix we provide numerical support for the claims of the paper title.

\begin{figure}[ht]
   \centering
   \includegraphics[width=0.9\linewidth]{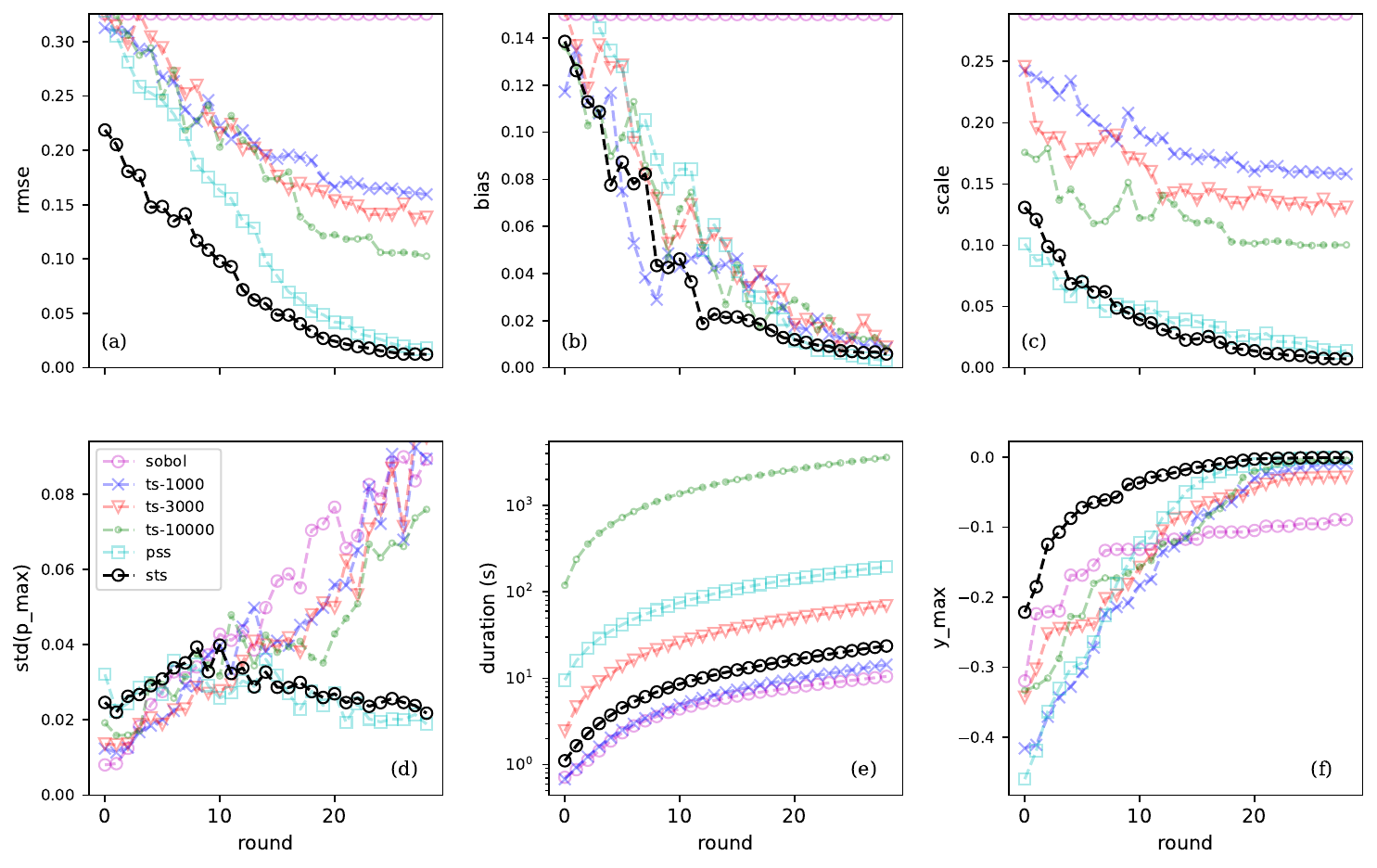}\hfill
    \caption{
    Comparison of STS to PSS and standard TS with varying numbers of candidates (1000, 3000, and 10,000). See  appendix~\ref{app:spt} for discussion. The optimizer, \texttt{sobol}, which proposes arms uniformly randomly, is included as a baseline.
    }
    \label{fig:tststs}
\end{figure}

Figure \ref{fig:tststs} compares STS to PSS and standard Thompson sampling using various numbers of candidate arms (100, 3000, and 10000). For each Thompson sampling method we maximized a sphere function
$$f(x) = -(x - 0.65\mathbf{1})^2$$
in the domain $x \in [0,1]^5$ where $\mathbf{1}$ is the vector of all 1's. At each round of the optimization we drew one arm, refit the GP, then drew 64 Thompson samples, $x_i$, solely for use in calculating summary statistics (i.e., not for use in the optimization). We have also included a uniformly-random sampler (\texttt{sobol}, not a Thompson sampler) for comparison. The samples were generated by a Sobol' quasi-random sampler \cite[Chapter~5]{dace}.

\textbf{Precision}: Subfigure (a) shows $rmse = \sum_i(x_i - 0.65)^2/64$, describing how near the Thompson samples are to the true optimum, $x=0.65\mathbf{1}$. Subfigures (b) and (c) decompose the RMSE into $bias = \sum_i(x_i - 0.65)/64$ and $scale = \big(\Pi_d s_d\big)^{1/5}$, where $s_d$ is the standard deviation of the $d^{\text{th}}$ dimension of the samples $x_i$. We note that the Thompson samples, $x_i$, get closer to the maximizer as (i) we increase the number of candidates in standard TS, (ii) the optimization progresses and more measurements are included in the GP, (iii) we switch from standard TS to PSS to STS. We note, also, that all Thompson samplers produce similarly unbiased samples, and that the improvement in RMSE comes from greater precision, i.e., reduced scale of the distribution of $x_i$.

\textbf{Thompson Sampling}: Next we support our claim that $x_i \sim p_*(x)$ by calculating $p_{\max,i}$, an estimate of the probability that sample $x_i$ is the maximizer over the 64 samples. To calculate $p_{\max,i}$ we take a joint sample $y_i \sim \mathcal{GP}(x_i)$ over the 64 $x_i$ and record which $x_i$ yields the largest $y_i$. We repeat this 1024 times and set $p_{\max,i} = \text{[count of times }x_i\text{ is the max]} / 1024$. The subfigure \texttt{std(p\_max)} shows the standard deviation of $p_{\max,i}$ over $i$. If all 64 samples $x_i$ were Thompson samples then we'd expect $p_{\max,i} = 1/64$ and $\text{std}(p_{\max})=0$. We see that $\text{std}(p_{\max})$ stays closer to zero for both PSS and STS, while the values for standard TS grow as the optimization progresses, similar to the uniformly-random sampler (\texttt{sobol}). [While low $\text{std}(p_{\max})$ is a necessary condition to claim that $x_i \sim p_*(x)$, it is not sufficient. For instance, there may be regions of $[0,1]^d$ where $p_*(x)>0$ but no $x_i$ appear.]

\textbf{Speed}: The running time (in seconds, subfigure (e)) is smaller for STS than for PSS or for standard TS with 10,000 candidates. Note that the y-axis has a logarithmic scale to show the separation between curves, although \texttt{duration} is linear in \texttt{round}. Subfigure (f) verifies that all methods optimize the sphere function. We configured PSS as in \cite{mtv}. (It is unclear whether the number of iterations used by the Hit-and-Run sampler was optimal or whether PSS could have been faster or slower if this number were tuned. In the next section, section~\ref{app:ablation}, we tune the number of iterations used by STS's Hit-and-Run and show that the value we used in the paper is optimal.)

Point (i), above, suggests that given enough candidates, TS might achieve the same small scale that STS does, although this would increase the running time of TS, and it is already much larger that of STS even at only 10,000 candidates.

\section{Ablation studies}
\label{app:ablation}

Stagger Thompson Sampling (STS), algorithm~\ref{alg:stagger}, modifies a Hit-and-Run sampler in two ways: (i) Instead of initializing $x_a$ randomly, it uses a guess at the maximizer, $\tilde{x}_* = \argmax_x \mu(x)$, and (ii) perturbation distances are drawn from a log-uniform distribution rather than uniformly.

\begin{figure}[ht]
   \centering
   \includegraphics[width=0.9\linewidth]{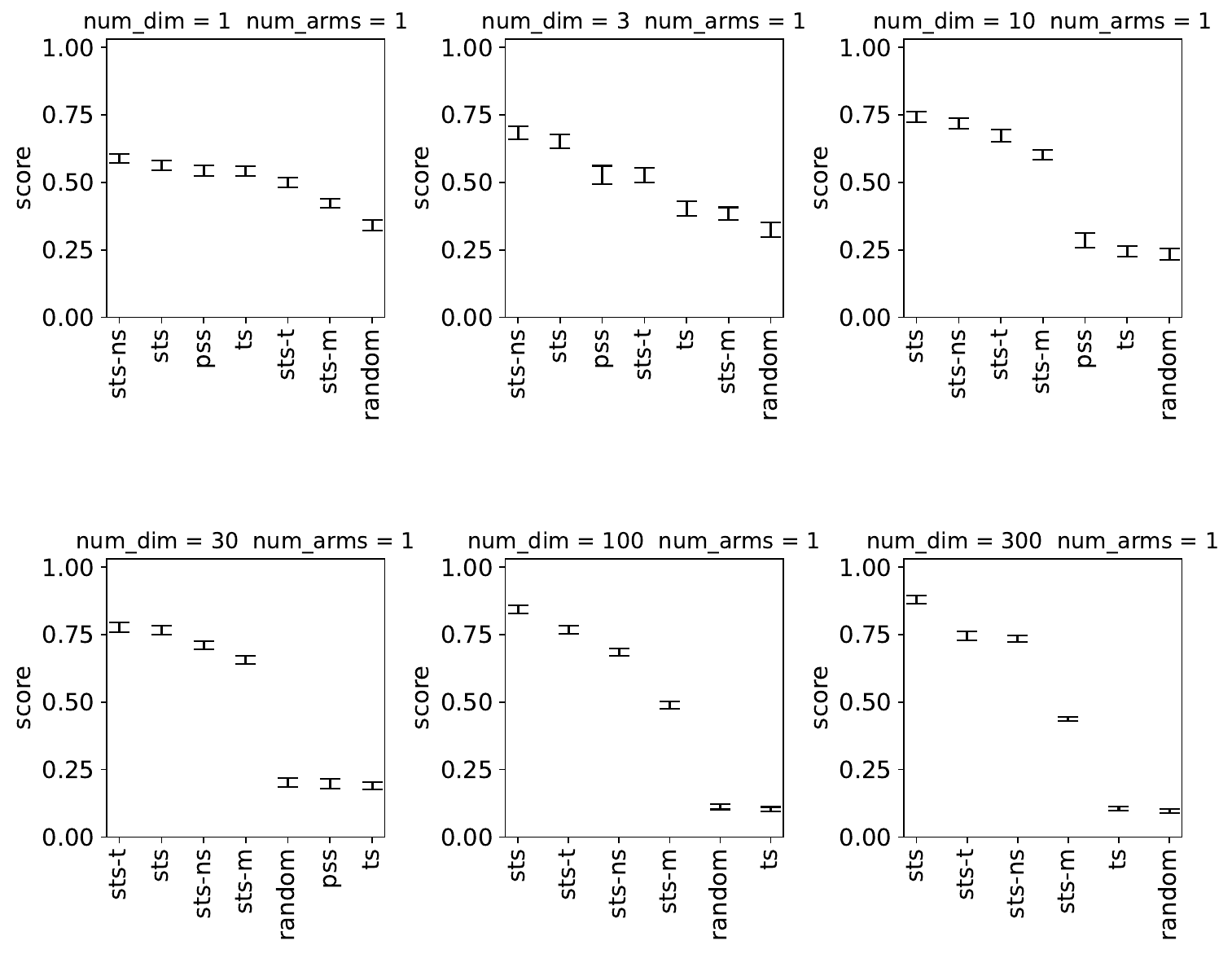}\hfill
    \caption{Ablations. \texttt{sts},  as presented in algorithm~\ref{alg:stagger},  performs as well as or better than any of the ablated version evaluated here. See text for descriptions of ablations..}
    \label{fig:ablate_algo}
\end{figure}

Figure~\ref{fig:ablate_algo} compares various ablations of STS:
\begin{itemize}
    \item \texttt{sts-ui} initializes $x_a$ uniformly-randomly
    \item \texttt{sts-m} initializes $x_a$ with the $x$ having the highest previously-measured $y$ value
    \item \texttt{sts-t} initializes $x_a$ to a Thompson sample from the $\mathcal{GP}$ at previously-measured $x$ values
    \item \texttt{sts-ns} replaces the stagger (log-uniform) perturbation with a uniform one
\end{itemize}
PSS, standard TS, and random arm selection are included for scale.
The figure shows that changing any of the features itemized above can reduce performance of STS.

\begin{figure}[ht]
   \centering
   \includegraphics[width=0.9\linewidth]{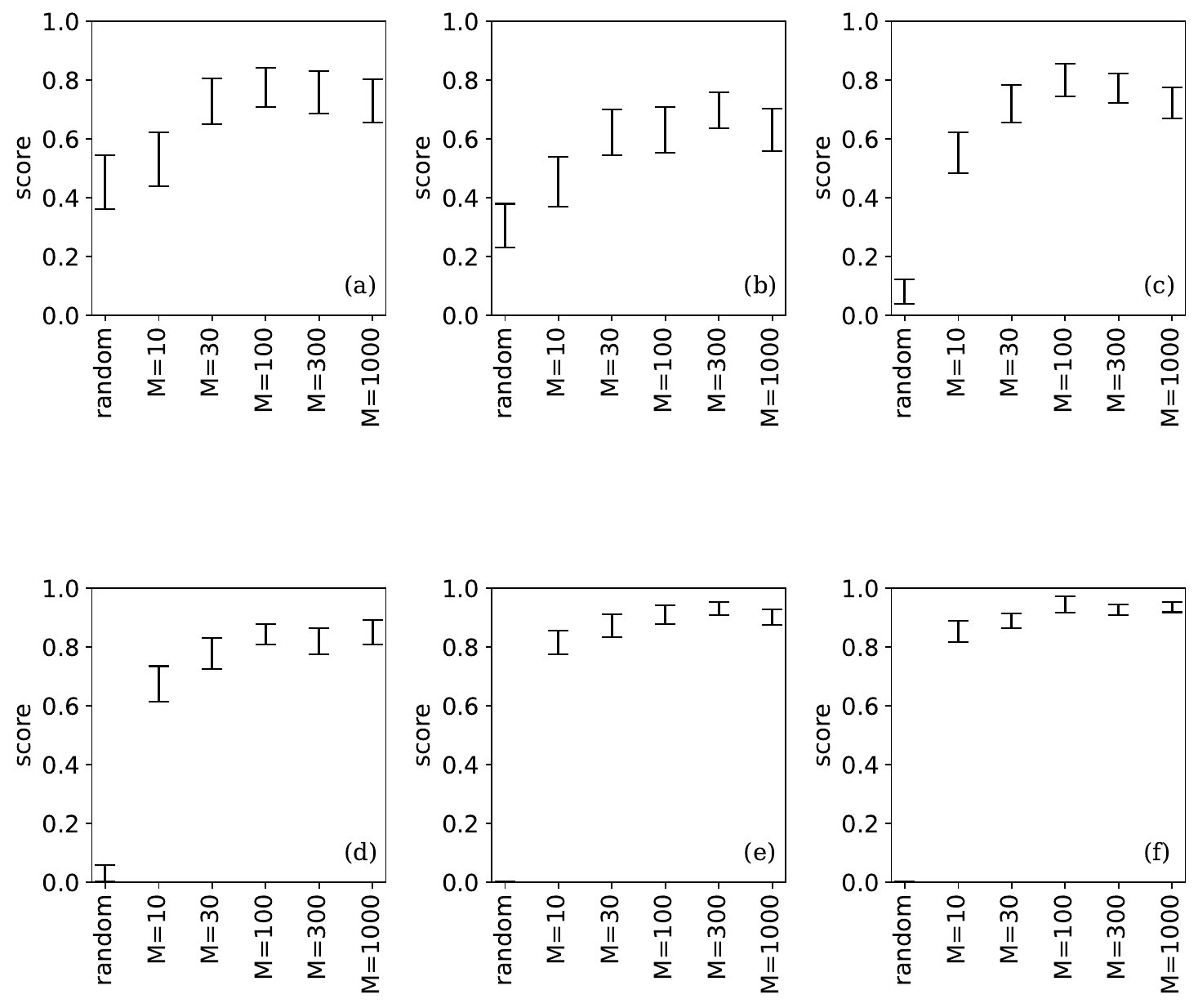}\hfill
    \caption{Performance stabilizes around $M=30$. Larger values of $M$ would increase running time for no meaningful benefit.}
    \label{fig:sweep}
\end{figure}

Figure~\ref{fig:sweep} sweeps values of a parameter, $M$, to STS, the number of iterations of the Hit-and-Run walk. See algorithm~\ref{alg:stagger} for details. The figure shows that performance asymptotes around $M=30$, which is the value used throughout the paper.

\end{document}